\documentstyle[12pt,fleqn]{article}
\newtheorem{theorem}{Theorem}[section]
\newtheorem{corollary}{Corollary}[section]
\newtheorem{lemma}{Lemma}[section]
\newtheorem{suggest}{Suggested Theorem}[section]
\newtheorem{definition}{Definition}[section]
\newtheorem{assumption}{Assumption}[section]

\newcommand{\xba}{\alpha}
\newcommand{\xbb}{\beta}

\newcommand{\xbe}{\in}
\newcommand{\xbf}{\phi}
\newcommand{\xbg}{\gamma}

\newcommand{\xbm}{\mu}

\newcommand{\xbq}{\psi}

\newcommand{\xCK}{\times}

\newcommand{\xCN}{\neg}
\newcommand{\xCQ}{\emptyset}

\newcommand{\xcc}{\subseteq}

\newcommand{\xce}{\not\in}

\newcommand{\xcm}{\models}

\newcommand{\xco}{\vee}
\newcommand{\xcp}{\rightarrow}

\newcommand{\xcr}{\leftrightarrow}
\newcommand{\xcs}{\cap}
\newcommand{\xcu}{\wedge}
\newcommand{\xcv}{\cup}

\newcommand{\xEd}{\neq}

\newcommand{\Xl}{\ldots}

\newcommand{\cL}{\mbox{${\cal L}$}}

\newcommand{\cU}{\mbox{${\cal U}$}}
\newcommand{\Cn}{\mbox{${\cal C}n$}}

\newcommand{\cP}{\mbox{${\cal P}$}}

\newcommand{\cZ}{\mbox{${\cal Z}$}}

\newcommand{\blackslug}{\mbox{\hskip 1pt \vrule width 4pt height 8pt
depth 1.5pt \hskip 1pt}}
\newcommand{\QED}{\quad\blackslug\lower 8.5pt\null\par\noindent}
\newcommand{\proof}{\par\penalty-1000\vskip .5 pt\noindent{\bf Proof\/: }}

\newcommand{\eqdef}{\stackrel{\rm def}{=}}
\newcommand{\lmid}{\mid\!}
\newcommand{\rmid}{\!\mid}
\title{
Preferred History Semantics for Iterated Updates
\thanks{This work was partially supported
by the Jean and Helene Alfassa fund for
research in Artificial Intelligence and by grant 136/94-1 of the
Israel Science Foundation on ``New Perspectives on Nonmonotonic Reasoning''.}
}
\author{
Shai Berger\thanks{Platonix Technologies Ltd., 44 Petach Tikva Road, Tel Aviv 
66183, Israel, shai@platonix.com}
\and
Daniel Lehmann\thanks{
Institute of Computer Science, Hebrew University, Jerusalem 91904, Israel,
lehmann@cs.huji.ac.il}
\and
Karl Schlechta\thanks{Laboratoire d'Informatique de Marseille, ESA CNRS 6077,
Universite de Provence, CMI, 39 rue Joliot-Curie, F-13453 Marseille
Cedex 13, France, ks@gyptis.univ-mrs.fr}
}

\date{01/3/99}

\begin{document}

\maketitle

\begin{abstract}

We give a semantics to iterated update by a preference relation on
possible developments. An iterated update is a sequence of formulas, giving
(incomplete) information about successive states of the world.
A development is a sequence of models, describing a possible
trajectory through time. We assume a principle of inertia
and prefer those developments, which are compatible with the
information, and avoid unnecessary changes.
The logical properties of the updates defined in this way are
considered, and a representation result is proved.

\end{abstract}

\section{Introduction}

\subsection{Overview}

We develop in this article an approach to update based on an abstract
distance
or ranking function. An agent has (incomplete, but reliable)
information (observations) about a changing
situation in the form of a sequence of formulas. At time 1, $ \xba_{1}$
holds, at time 2,
$ \xba_{2}$ holds, $ \Xl.,$ at time $n$, $ \xba_{n}$ holds. We are thus in
a situation of iterated
update. The agent tries to reason about the most likely
outcome, i.e. to sharpen the information $ \xba_{n}$ by plausible
reasoning. He knows that
the real world has taken some trajectory, or history, that can be
described by
a sequence of models $<m_{1}, \Xl.,m_{n}>,$ where $m_{i} \xcm \xba_{i}$
(remember the observations
were supposed to be reliable). We say that such a history explains
the observations. For his reasoning, he makes two assumptions:
First, an assumption of inertia: histories that stay constant
are more
likely than histories that change without necessity. For instance, if
$n=2,$
and $ \xba_{1}$ is consistent with $ \xba_{2},$ $m_{1} \xcm \xba_{1} \xcu
\xba_{2},$ $m_{2} \xcm \xba_{2},$ then the history $\langle m_{1},m_{1}\rangle$
is preferred to the history $\langle m_{1},m_{2}\rangle.$ We do NOT assume that
$\langle m_{1},m_{1}\rangle$ is more
likely than some $\langle m_{3},m_{2}\rangle,$ i.e. we do not compare the 
cardinality
of changes,
we only assume ``sub-histories''  to be more likely than longer ones.
Second, the agent assumes that histories can be ranked by their
likelihood, i.e.
that there is an (abstract) scale, a total order, which describes this
ranking.
These assumptions are formalized in Section 1.4.

The agent then considers those models of $ \xba_{n}$ as most plausible,
which are
endpoints of preferred histories explaining the observations. Thus, his
reasoning defines an operator $[\;]$ from the set of sequences of
observations
to the set of formulas of the underlying language, s.t. $[ \xba_{1}, \Xl
., \xba_{n}] \xcm \xba_{n}.$

The purpose of this article is to characterize the operators $[\;]$ that
correspond to such reasoning.
Thus, we will give conditions for the operator $[\;]$ which all operators
based on history ranking satisfy, and, conversely, which allow to construct
a ranking $r$ from the operator $[\;]$  such that the operator $[\;]_r$
based on this ranking is exactly $[\;]$. The first part can be called
the soundness, the second the completeness part.

Before giving a complete set of conditions in Section 3, we discuss in
Section 2 some logical and intuitive properties of ranking based
operators, in particular those properties which are related to the 
Alchourr\'on, G\"{a}rdenfors, Makinson postulates for theory revision, or to 
the update postulates of Katsuno, Mendelzon. In Section 3, we give a full 
characterization of these operators. We start from a result of Lehmann, 
Magidor, Schlechta ~\cite{LMS:98} on distance based revision, which has some 
formal similarity, and refine the techniques developed there.

In the rest of this section, we first compare briefly revision and update,
and then emphasize the relevance of epistemic states for iterated update,
i.e. that belief sets are in general insufficient to determine the
outcome of our reasoning about iterated update. We then make our approach
precise, give some basic definitions, and recall the AGM and KM
postulates.

\subsection{Revision and Update}

Intuitively, belief revision (also called theory revision)
deals with evolving knowledge about a static situation.
Update, on the other hand, deals with knowledge about an evolving
situation. It is not clear that this ontological distinction agrees
with the semantical and proof-theoretic distinction between the
AGM and the KM approaches. In this paper, the distinction
between revision and update must be understood as ontological,
not as AGM vs. KM semantics or postulates .

In the case of belief revision, an agent receives successively
different information about a situation,
e.g. from different sources, and the union of this information may
be inconsistent. The theory of belief revision describes ``rational" ways to
incorporate new information into a body of old information, especially when
the new and old information together are inconsistent.

In the case of update, an agent is informed that at time $t$, a formula $\phi$ 
held, at
time $t'$ $\phi'$, etc. The agent tries, given this information and some
background assumptions (e.g. of inertia: that things do not change
unless forced to do so) to draw plausible conclusions about
the probable development of the situation.
This distinction goes back to Katsuno and Mendelzon~\cite{KatMend:92}.

Revision in this sense is formalized in
Lehmann, Magidor and Schlechta~\cite{SLM:96}, elaborated in~\cite{LMS:98}.
The authors have devised there a
family of semantics for revision based on minimal change,
where change is measured by distances between models of formulae of the
background logic. More precisely,
the operator defined by such functions revises a belief set $T$ according to
a new observation $\xba$ by picking the models of $\xba$ that are closest to
models of $T$.
Such revision operators have been shown to satisfy some of the common
rationality postulates. Several weak forms of a distance
function (``pseudo-distances") have been studied in ~\cite{LMS:98},
and representation theorems for abstract revision
operators by pseudo-distances have been proved.
Since, in the weak forms, none of the notions usually connected with a distance
(e.g. symmetry, the triangle inequality) are used, such pseudo-distances are
actually no more than a preference function, or a ranked order, over pairs of
models.

In the present article, we consider a setting that intuitively has an ontology 
of update. It sets a single belief change system for all sequences of 
observations. All pseudo-distances are between individual models,
as in the semantics proposed in~\cite{KatMend:92}.
But, where Katsuno and Mendelzon's semantics takes a ``local'' approach and 
incorporates in the new belief set the best updating models for each model in 
the old belief set, we take a ``global'' approach and pick for the updated 
belief set only the ending models of the best overall histories. As a result, 
our system validates all the AGM postulates, and not all the KM ones. The 
introduction of an update system which follows the AGM postulates for revision 
may have some interesting ontological consequences, but these will not be dealt 
with in this work.

The formal definitions and assumptions are to be found in Definitions
~\ref{def:explains}, ~\ref{def:preferred}, 1.3, and Assumption 1.1.
We prove a representation theorem for update
operators based on rankings of histories,
similar to those in ~\cite{LMS:98}.
Note that the approach taken here is more specific than that of
\cite{Sch:95}, which considers arbitrary (e.g. not necessarily
ranked) preference relations between histories.

\subsection{Epistemic States are not Belief Sets}

The epistemic state of an agent, i.e., the state of its mind, is understood
to include anything that influences its actions, its beliefs about what
is true of the world, and the way
it will update or revise those beliefs, depending on the information it
gathers.
The belief set of an agent, at any time, includes only the set of propositions
it believes to be true about the world at this time.
One of the components of epistemic states must therefore be the belief set
of the agent.
One of the basic assumptions of the AGM theory of belief revisions
is that epistemic states are belief sets, i.e., they do not include
any other information.
At least, AGM do not formalize in their basic theory, as expressed by
the AGM postulates, any incorporation of other information in the
belief revision process.
In particular an agent that holds exactly the same beliefs about
the state of the world, at two different
instants in time is, at those times, in the same epistemic state and therefore,
if faced with the same information, will revise its beliefs in the same way.
Recent work on belief revision and update has shown this assumption
has very powerful consequences, not always
welcome~\cite{Leh:IJCAI95,ADJP:96,NFJH:96}.
Earlier work on belief base revision (see e.g.~\cite{Nebel}) expresses a 
similar concern about the fundamentals of belief revision.

We do not wish to take a stand on the question of whether this identification
of epistemic states with belief sets is reasonable for the study of belief
revision, but we want to point out that, in the study of belief update,
with its natural sensitivity (by the principle of inertia) to the order in 
which the information is gathered, it is certainly unreasonable.
This is illustrated by the following observation:
Let $\xbf$ and $\xbq$ be different atomic, i.e. logically independent, 
formulas.
First scenario: update the trivial belief set (the set of all
tautologies)
by $\xbf$, then by $\xbq$, then by \mbox{$\neg \xbf \vee \neg \xbq$}.
Second scenario: update the trivial belief set by $\xbq$, then by $\xbf$, then
by
\mbox{$\neg \xbf \vee \neg \xbq$}.
We expect different belief sets: we shall most probably try to stick to the
piece of information that is the most up-to-date, i.e., $\xbq$
in the first scenario and $\xbf$ in the second scenario.
But there is no reason for us to think that the belief sets
obtained, in both scenarios, just before the last updates, should be different.
We expect them to be identical: the consequences of \mbox{$\xbf \wedge \xbq$}.
The same agent, in two different epistemic states, updates differently
the same beliefs in the light of the same information.

\subsection{Preferred History Semantics}

We now make our approach more precise.
The basic ontology is minimal: The agent makes a sequence of observations and
interprets this sequence of observations in terms of possible
histories of the world explaining the observations.

Assume a set $\cL$ of formulas and a set $\cU$ of models for $\cL$.
Formulas will be denoted by Greek letters from the beginning of the alphabet:
$\xba$, $\xbb$ and so on, and models by $m$, $n$, and so on.
We do not assume formulas are indexed by time.
An observation is a {\em consistent} formula. (We assume observations to be
consistent for two reasons: First, observations are assumed to be reliable;
second, as we work with histories made of models, we need some model to explain
every observation. Working with unreliable information would be the subject of
another paper.) Observing a formula means observing the formula holds.

A sequence of observations is here a $finite$ sequence of observations.
Sequences of observations will be denoted by Greek letters from the
end of the alphabet:
$\sigma$, $\tau$ and concatenation of such sequences by $\cdot$.
Notice the empty sequence is a legal sequence of observations.
We shall identify an observation with the sequence of observations
of length one that contains it.
What does a sequence of observations tell us about the present
state of the world?

A history is a finite, non-empty sequence of models.
Histories will be denoted by $h$, $f$, and so on.

\begin{definition}
\label{def:explains}
A history \mbox{$h = \langle m_{0} , \ldots , m_{n} \rangle$}
{\em explains} a sequence of observations
\mbox{$\tau = \langle \xba_{0} , \ldots , \xba_{k} \rangle$}
iff there are subscripts
\mbox{$0 \leq i_{0} \leq i_{1} \leq \ldots \leq i_{k} \leq n$}
such that for any $j$, \mbox{$0 \leq j \leq k$},
\mbox{$m_{i_{j}} \models \xba_{j}$}.
\end{definition}

Thus, a history explains a sequence of observations
if there is, in the history, a model that explains each of the observations
in the correct order.
Notice that $n$ is in general different from $k$, that
many consecutive $i_{j}$'s may be equal, i.e., the same model
may explain many consecutive observations,
that some models of the history may not be used at all in the explanation,
i.e., $l$, \mbox{$0 \leq l \leq n$} may be equal to none of the
$i_{j}$'s,
and that we do not require
that $j_{k}$ be equal to $n$, or that $j_{0}$ be equal to $0$,
i.e., there may be useless models even at the start or at the end of a history.
Note also that if $h$ explains a sequence $\sigma$ of observations, it
also explains any subsequence (not necessarily contiguous) of $\sigma$.

The set
of histories that explain a sequence of observations give us information
about the probable outcome.
Monotonic logic is useless here: if we consider all histories explaining
a sequence of observations, we cannot conclude anything.
It is reasonable, therefore to assume the agent restricts the
set of histories it considers to a subset of the explaining histories.

We shall assume the agent has some preferences among histories,
some histories being more natural, simpler, more expected, than others.
A sequence of observations defines thus a subset of the set of all histories
that explain it: the set of all preferred histories that explain it.
This set defines the set of beliefs that result
from a sequence of observations: the set of formulas satisfied in all the
models that may appear as last elements of a preferred history that
explains the sequence.
The beliefs held depend on the preferences, concerning histories,
of the agent.
The logical properties of update depend on the class of preferences
we shall consider.

Formally, one assumes the agent's preferences are represented by a
binary relation $<$ on histories.
Intuitively, \mbox{$h < f$} means that history $h$ is strictly
preferred, e.g. strictly more natural or strictly simpler, than
history $f$.
Note that our relation is on histories, not on models.
We may now define preferred histories.

\begin{definition}
\label{def:preferred}
A history $h$ is a preferred history for a sequence $\sigma$ of observations
iff

\begin{itemize}
\item $h$ explains $\sigma$
\item there is no history $h'$ that explains $\sigma$ such that
\mbox{$h' < h$}.
\end{itemize}

\end{definition}

In this work two assumptions are made concerning the preference relation $<$.
First, we assume $<$ is a strict modular, well-ordering, i.e.,
$<$ is irreflexive, transitive, if \mbox{$h < h'$}, then for any
$f$, either \mbox{$h < f$} or \mbox{$f < h'$} and there is no infinite
descending chain.
Secondly, we assume that partial histories (i.e., sub-histories)
are preferred over (longer) histories:

\begin{assumption}
\label{ass:preferred}
If \mbox{$h = \langle m_{0} , \ldots , m_{n} \rangle$} and
\mbox{$h' = \langle m_{j_{0}} , \ldots , m_{j_{k}}\rangle$}, for
\mbox{$0 \leq j_{0} < j_{1} < \ldots < j_{k} \leq n$} and
\mbox{$0 \leq k < n$}, then \mbox{$h' < h$}.
\end{assumption}

For instance, $h'=\langle m_2 , m_4\rangle$ is preferred to
$h=\langle m_1, m_2 , m_3, m_4\rangle$.

This assumption is justified both by an epistemological concern:
simpler explanations are better, and by an assumption of inertia
concerning the way the universe evolves: things tend to stay as they are. This
assumption, in a finite setting, trivializes the well-ordering assumption.

Finally, we formally define our operator $[\;]$:

\begin{definition}
\label{def:beliefset}
After a sequence $\sigma$ of observations, the agent holds the
beliefs \mbox{$[ \sigma ]$}, defined by
\mbox{$\xba \in [ \sigma ]$} iff for every model $m$ and every history
$h$, if $h$ is a preferred history explaining $\sigma$ and $m$ is the last
element of $h$, then \mbox{$m \models \xba$}.
\end{definition}

Notice that, since histories are non-empty, this definition always
makes sense.

The remainder of this paper will show that the assumptions above have
far reaching consequences: they are strong assumptions.
Our purpose is indeed to look for a powerful logic, not for the minimal
logic that agrees with any possible ontology.

\subsection{Basic Definitions and Notation}

We are dealing here only with finite and complete
universes, so theories have logically equivalent formulas (and vice versa),
and are isomorphic to sets of models, and we will use them in these senses 
interchangeably. We will see sequence concatenation also as an outer product 
(with respect to concatenation) of sets of histories. For technical reasons, 
most of the
discussion will relate to sets of models. An exception, for easier readability
and comparison with the AGM and KM conditions is Section 2.
We do
not consistently differentiate singletons from the members that comprise
them, because it is always clear from context which of them we are referring
to.

A theory, or belief set, will be a deductively closed set of formulas.

For the convenience of the reader, we recall the AGM postulates for belief
revision, (see e.g. ~\cite{Gard:intro}),
and the Katsuno-Mendelzon postulates for update
(see e.g.
~\cite{KatMend:92}):

\vspace{0.2cm}

$(K*1)$ $K* \xba $ is a deductively closed set of formulas.

$(K*2)$ $ \xba \xbe K* \xba.$

$(K*3)$ $K* \xba \xcc Cn(K, \xba ).$

$(K*4)$ If $ \xCN \xba \xce K,$ then $Cn(K, \xba ) \xcc K* \xba.$

$(K*5)$ If $K* \xba $ is inconsistent then $ \xba $ is a logical
contradiction.

$(K*6)$ If $ \xcm \xba \xcr \xbb $, then $K* \xba =K* \xbb.$

$(K*7)$ $K* \xba \xcu \xbb \xcc Cn(K* \xba, \xbb ).$

$(K*8)$ If $ \xCN \xbb \xce K* \xba,$ then $Cn(K* \xba, \xbb ) \xcc K*
\xba \xcu \xbb.$

\vspace{0.2cm}

(U1) $ \xcm ( \xbq  \cdot \xbm ) \xcp \xbm.$

(U2) If $ \xcm \xbq \xcp \xbm,$ then $ \xcm ( \xbq  \cdot \xbm ) \xcr \xbq.$

(U3) If both $ \xbq $ and $ \xbm $ are satisfiable, then so is $ \xbq  \cdot
\xbm.$

(U4) If $ \xcm \xbq_{1} \xcr \xbq_{2}$ and $ \xcm \xbm_{1} \xcr \xbm_{2},$
then $ \xcm ( \xbq_{1} \cdot \xbm_{1}) \xcr ( \xbq_{2} \cdot \xbm_{2}).$

(U5) $ \xcm (( \xbq  \cdot \xbm ) \xcu \xbf ) \xcp ( \xbq  \cdot( \xbm \xcu
\xbf
)).$

(U6) If $ \xcm ( \xbq  \cdot \xbm_{1}) \xcp \xbm_{2}$ and $ \xcm ( \xbq  \cdot
\xbm_{2}) \xcp \xbm_{1},$ then $ \xcm ( \xbq  \cdot \xbm_{1}) \xcr ( \xbq
\cdot
\xbm_{2}).$

(U7) If $ \xbq $ is complete, then $ \xcm ( \xbq  \cdot \xbm_{1}) \xcu ( \xbq
\cdot
\xbm_{2}) \xcp ( \xbq  \cdot( \xbm_{1} \xco \xbm_{2})).$

(U8) $ \xcm (( \xbq_{1} \xco \xbq_{2}) \cdot \xbm ) \xcr ( \xbq_{1} \cdot \xbm
)
\xco ( \xbq_{2} \cdot \xbm ).$

\vspace{0.2cm}

The following is a slight reformulation of the AGM postulates
in the spirit of Katsuno-Mendelzon, taken from
~\cite{LMS:98}. We consider here a symmetrical version, in the sense that
$K$ and $\alpha$ can both be theories, and simplify by considering only
consistent theories.

\vspace{0.2cm}

$(*0)$ If $ \xcm T \xcr S,$ $ \xcm T' \xcr S',$ then $T*T' =S*S'
,$

$(*1)$ $T*T' $ is a consistent, deductively closed theory,

$(*2)$ $T' \xcc T*T',$

$(*3)$ If $T \xcv T' $ is consistent, then $T*T' =Cn(T \xcv T' ),$

$(*4)$ If $T*T' $ is consistent with $T'',$ then $T*(T' \xcv
T'' )=Cn((T*T' ) \xcv T'' ).$

\vspace{0.2cm}

Finally, we recall the definition of a pseudo-distance from
~\cite{LMS:98}.

\begin{definition}

$\hspace{0.5em}$

$d:U \xCK U \xcp Z$ is called a pseudo-distance on $U$ iff
$Z$ is totally ordered by a relation $<.$

\end{definition}

\section{Some Important Logical Properties of Updates}

A number of logical properties of the operator $[ \; ]$ will
now be described and discussed.
The reasons why those properties hold are varied: some depend on very little
of our assumptions, some on almost all of them.
We shall try to make the appropriate distinctions.

\begin{lemma}
\label{le:theory}
For any $\sigma$, \mbox{$[ \sigma ]$} is a theory.
\end{lemma}

This property is analogous to AGM's $(K*1)$
and is implicit in Katsuno-Mendelzon's presentation~\cite{KatMend:92}.
This depends only on the fact that Definition~\ref{def:beliefset}
defines \mbox{$[ \sigma ]$} as the set of all formulas that hold
for all the models in a given set.
Indeed, this is a property that is expected to hold by the
structure of belief sets, not by the definition of explanation
or certain properties of the preference relation.

The following properties hold by the definition of explanation,
i.e., Definition~\ref{def:explains}.

\begin{lemma}
\label{le:equiv}
If $\xba$ and $\xba'$ are logically equivalent, then for any sequences
$\sigma$, $\tau$:
\mbox{$[ \sigma \cdot \xba \cdot \tau ] = [ \sigma \cdot \xba' \cdot \tau ]$}.
\end{lemma}

This property is analogous to AGM's $(K*6)$
but notice that, there, it is needed only for the second argument
of the revision operation, since it is implicit for the first,
a theory.
It parallels (U4) in Katsuno-Mendelzon's~\cite{KatMend:92}.
Lemma~\ref{le:equiv} follows from Definition~\ref{def:explains},
that implies that the histories that explain
\mbox{$ \sigma \cdot \xba \cdot \tau $} are exactly those that explain
\mbox{$ \sigma \cdot \xba' \cdot \tau $}.
The preferred histories are therefore the same.
The next property is more original.

\begin{lemma}
\label{le:strong}
If \mbox{$\xbb \models \xba$}, then for any sequences
$\sigma$, $\tau$:
\[
[ \sigma \cdot \xba \cdot \xbb \cdot \tau ] =
[ \sigma \cdot \xbb \cdot \tau ] =
[ \sigma \cdot \xbb \cdot \xba \cdot \tau ].
\]
\end{lemma}

This property has no clear analogue in the AGM or KM frameworks,
but is closely related to (U2) of~\cite{KatMend:92}.
The first equation is property (C1) of Darwiche-Pearl's~\cite{ADJP:96}
and (I5) of~\cite{Leh:IJCAI95}.
The second equation is a weakening of (I4) of~\cite{Leh:IJCAI95}.
Here we request $\xba$ to be a logical consequence of $\xbb$, there
we only asked that $\xba$ be in \mbox{$[\sigma \cdot \xbb ]$}.
Lemma~\ref{le:strong} is a consequence of the fact that
the histories that explain \mbox{$ \sigma \cdot \xbb \cdot \tau $},
\mbox{$ \sigma \cdot \xba \cdot \xbb \cdot \tau $} and
\mbox{$ \sigma \cdot \xbb \cdot \xba \cdot \tau $} are the same.

\begin{corollary}
\label{co:true}
For any sequence $\sigma$,
\mbox{$[ \sigma \cdot {\bf true} ] = [ \sigma ]$}.
\end{corollary}

The next property deals with disjunction.

\begin{lemma}
\label{le:disj1}
If $\xbg$ is a member both of \mbox{$ [ \sigma \cdot \xba \cdot \tau ] $}
and \mbox{$ [ \sigma \cdot \xbb \cdot \tau ] $}, then it is a member of
\mbox{$ [ \sigma \cdot \xba \vee \xbb \cdot \tau ] $}.
In other words
\[
[ \sigma \cdot \xba \cdot \tau ] \cap [ \sigma \cdot \xbb \cdot \tau ]
\subseteq [ \sigma \cdot \xba \vee \xbb \cdot \tau ].
\]
\end{lemma}

This property is similar to one half of (U8) of~\cite{KatMend:92}, and
to a consequence of AGM's (K*7), as pointed out in~\cite{Gard:book}, property
(3.14): $(K*A) \xcs (K*B) \xcc K*(A \xco B)$.

Lemma~\ref{le:disj1} depends only on Definitions~\ref{def:explains}
and~\ref{def:preferred}, but does not depend on any properties
of the preference relation.
A history $h$ that explains \mbox{$ \sigma \cdot \xba \vee \xbb \cdot \tau$}
explains \mbox{$ \sigma \cdot \xba \cdot \tau$} or
\mbox{$ \sigma \cdot \xbb \cdot \tau$}.
A preferred history for \mbox{$ \sigma \cdot \xba \vee \xbb \cdot \tau$}
must therefore either be a preferred history for
\mbox{$ \sigma \cdot \xba \cdot \tau$} (since any history explaining the latter
explains the former)
or  preferred history for \mbox{$ \sigma \cdot \xbb \cdot \tau$}.
The next lemma is a strengthening of Lemma~\ref{le:disj1},
and it depends on the modularity of the preference relation.

\begin{lemma}
\label{le:disj2}
The theory \mbox{$ [ \sigma \cdot \xba \vee \xbb \cdot \tau ] $} is equal
to \mbox{$ [ \sigma \cdot \xba \cdot \tau ] $}, equal to
\mbox{$ [ \sigma \cdot \xbb \cdot \tau ] $} or is the intersection of the two
theories above.
\end{lemma}

This property is a weakening of (U8) of~\cite{KatMend:92}, compare also to
property (3.16) in~\cite{Gard:book}:
$K*(A \xco B)=K*A$ or $K*(A \xco B)=K*B$ or $K*(A \xco B)=(K*A) \xcs (K*B)$.

\proof
A history $h$ that explains \mbox{$ \sigma \cdot \xba \vee \xbb \cdot \tau$}
explains \mbox{$ \sigma \cdot \xba \cdot \tau$} or
\mbox{$ \sigma \cdot \xbb \cdot \tau$}.
If all preferred histories for \mbox{$ \sigma \cdot \xba \vee \xbb \cdot \tau$}
explain \mbox{$ \sigma \cdot \xba \cdot \tau$}, then

\begin{itemize}
\item any preferred history for \mbox{$ \sigma \cdot \xba \vee \xbb \cdot
\tau$}
is a preferred history for \mbox{$ \sigma \cdot \xba \cdot \tau$}
(otherwise there would be a strictly preferred history for
\mbox{$ \sigma \cdot \xba \cdot \tau$}, but that explains
\mbox{$ \sigma \cdot \xba \vee \xbb \cdot \tau$}), and
\item any preferred history for \mbox{$ \sigma \cdot \xba \cdot \tau$}
is a preferred history for \mbox{$ \sigma \cdot \xba \vee \xbb \cdot \tau$},
otherwise there would be a strictly preferred history for
\mbox{$ \sigma \cdot \xba \vee \xbb \cdot \tau$} that satisfies
\mbox{$ \sigma \cdot \xbb \cdot \tau$}.
\end{itemize}

In this case, \mbox{$ [ \sigma \cdot \xba \vee \xbb \cdot \tau ] $} is equal
to \mbox{$ [ \sigma \cdot \xba \cdot \tau ] $}.

Similarly, if all preferred histories for
\mbox{$ \sigma \cdot \xba \vee \xbb \cdot \tau$}
explain \mbox{$ \sigma \cdot \xbb \cdot \tau$}, then
\mbox{$ [ \sigma \cdot \xba \vee \xbb \cdot \tau ] $} is equal
to \mbox{$ [ \sigma \cdot \xbb \cdot \tau ] $}.

Let us assume, therefore, that some preferred histories for
\mbox{$ \sigma \cdot \xba \vee \xbb \cdot \tau$} explain
\mbox{$ \sigma \cdot \xba \cdot \tau$} and that some explain
\mbox{$ \sigma \cdot \xbb \cdot \tau$}.
By modularity of the preference relation, any preferred history
for \mbox{$ \sigma \cdot \xba \cdot \tau$} is a preferred history for
\mbox{$ \sigma \cdot \xba \vee \xbb \cdot \tau$}.
\QED
The next properties follow from the assumption that sub-histories
are preferred to more complete histories.
Remark first that, if a history $h$
explains a non-empty sequence $\sigma$ of observations
but the last model of $h$
does not satisfy the last observation of $\sigma$, then there is
a shorter (initial) sub-history of $h$ that explains $\sigma$.
The history $h$ cannot be, in this case, a preferred history for $\sigma$.

\begin{lemma}
\label{le:K2}
For any sequence $\sigma$ of observations and any formula $\xba$:
\mbox{$\xba \in [ \sigma \cdot \xba ] $}.
\end{lemma}

This property is similar to AGM's $(K*2)$
and (U1) of~\cite{KatMend:92}.
In~\cite{NFJH:96}, Friedman and Halpern question this postulate.
Here, it finds a justification, grounded in our preference for
shorter explanations.

\begin{lemma}
\label{le:K7K8}
For any sequence $\sigma$ of observations and any formulas
$\xba$ and $\xbb$, if \mbox{$\neg \xbb \not \in [ \sigma \cdot \xba ] $},
then \mbox{$[ \sigma \cdot \xba \cdot \xbb ] = [ \sigma \cdot \xba \wedge \xbb
]
= \Cn([\sigma \cdot \xba] , \xbb)$}.
\end{lemma}

This property is analogous to AGM's $(K*7)$ and
$(K*8)$.
\proof
We show that the preferred histories for \mbox{$\sigma \cdot \xba \cdot \xbb$}
are exactly those preferred histories of \mbox{$\sigma \cdot \xba$} whose
last element satisfies $\xbb$.
First, clearly, any preferred history for \mbox{$\sigma \cdot \xba$}
whose last element satisfies $\xbb$ explains \mbox{$\sigma \cdot \xba \cdot
\xbb$}
and is a preferred history for it.
Secondly, since \mbox{$\neg \xbb \not \in [ \sigma \cdot \xba ] $},
there is a preferred history $h$ for
\mbox{$\sigma \cdot \xba$} whose last element satisfies $\xbb$.
As we have just seen $h$ is a preferred history for
\mbox{$\sigma \cdot \xba \cdot \xbb$}.
Let $f$ be a preferred history for \mbox{$\sigma \cdot \xba \cdot \xbb$}.
It explains \mbox{$\sigma \cdot \xba$}. If it were not a preferred
history for \mbox{$\sigma \cdot \xba$}, there would be a history
$f'$, \mbox{$f' < f$} that explains \mbox{$\sigma \cdot \xba$}.
By modularity, we would have \mbox{$f' < h$} or \mbox{$h < f$}, which are both
impossible.
We conclude that $f$ is a preferred history for \mbox{$\sigma \cdot \xba$}.
But its last element satisfies $\xbb$, by Lemma~\ref{le:K2}.

We have shown that
\mbox{$[ \sigma \cdot \xba \cdot \xbb ] = \Cn([\sigma \cdot \xba] , \xbb)$}.

To conclude the proof, notice that, by the above, any preferred history
for \mbox{$ \sigma \cdot \xba \cdot \xbb$} explains \mbox{$\sigma \cdot \xba
\wedge \xbb$}
and that any history explaining
\mbox{$\sigma \cdot \xba \wedge \xbb$}
also explains \mbox{$ \sigma \cdot \xba \cdot \xbb$}.
\QED

\begin{corollary}
\label{co:K3K4}
If \mbox{$ \neg \xba \not \in [\sigma]$}, then
\mbox{$[\sigma \cdot \xba] = \Cn([\sigma] , \xba)$}.
\end{corollary}

This parallels AGM's $(K*3)$ and $(K*4)$.
\proof
By Corollary~\ref{co:true},
\mbox{$[\sigma] = [\sigma \cdot {\bf true}]$}.
By Lemma~\ref{le:K7K8},
\[
[\sigma \cdot {\bf true} \cdot \xba] =
\Cn([\sigma \cdot {\bf true}] , \xba) =
[\sigma \cdot {\bf true} \wedge \xba].
\]
By Corollary~\ref{co:true} and  Lemma~\ref{le:strong}
\mbox{$\Cn([\sigma] , \xba) = [\sigma \cdot \xba]$}.
\QED
Our last property depends on well-foundedness, which is trivial in a
finite setting.

\begin{lemma}
\label{le:cons}
For any sequence $\sigma$, \mbox{$[ \sigma ]$} is consistent.
\end{lemma}

This parallels AGM's $(K*5)$ and KM's (U3).
This property depends on two assumptions.
First we assumed observations were consistent formulas.
It follows that any sequence of observations is explained by some
history.
By finiteness, if $h$ explains $\sigma$ and $h$ is not a preferred
history for $\sigma$, then there is a preferred history $h'$ for $\sigma$
(in fact one such that \mbox{$h' < h$}).
Therefore \mbox{$[ \sigma ]$} is consistent.

\section{A Representation Theorem}

\subsection{Introduction}

In Section 2, we have presented several logical properties of operators $[\;]$
based on history ranking. In this Section, we will give a full 
characterization. We generalize here results about revision reported in 
~\cite{LMS:98}. We will first show that a straightforward generalization of the 
not necessarily symmetrical revision case fails already for sequences of length 
3, and will then give a characterization for sequences of arbitrary finite 
length in Theorem 3.2.
A technical problem for the latter is that we have to work with ``illegal''
sets of histories, which do not correspond to sequences of sets of models. E.g.
the set of sequences $\{\langle 0,0,0 \rangle, \langle 0,1,1 \rangle\}$ is
not the product of any sequence of sets - $\{0\} \times \{0,1\} \times
\{0,1\}$ contains too many sequences. Our operator is, however, only defined 
for such sequences of sets. We use the idea of a patch, a cover of such sets of
sequences by products of sequences of sets, to show our result.

As mentioned before, we will want to work mainly with sequences of sets of
models. Such sets are freely interchangeable with observations, and every
sequence thereof also defines a set of histories. We would like the set of all
explaining histories to be representable as a sequence of sets of models.
Definition~\ref{def:explains} allows the set of explaining histories to be
infinite, and we will limit the sets of histories dealt with by an assumption
strengthening Lemma~\ref{le:K2}.
While we do not assume full sub-history preference, we assume that a history
explains a sequence of observations if they are of the same length, and the
$i$th model in the history models the $i$th observation in the sequence. This
assumption, like the Lemma, is justified by sub-history preference, with an
implicit agreement that consecutive repeats of a model in a history are merely
another way to write that the model explains several observations. In other
words, to make the formal phrasing and proof a little easier, we will write for
each observation in the sequence $\sigma$ the model that explains it in the
history $h$. A history $h'$ containing $h$ as a sub-history will not be
preferable to $h$, and sub-histories of $h$ are just represented as longer than
they are. Intuitively, an assumption is made here that e.g. \mbox{$\langle
m_1,m_1,m_2 \rangle $} and \mbox{$\langle m_1,m_2,m_2 \rangle $}, both being
representations of \mbox{$\langle m_1,m_2 \rangle $}, are equally preferred,
and are both considered better than \mbox{$\langle m_1,m_2,m_3 \rangle $}. This
assumption is neither used nor needed in the theorem or its proof. We make no
further assumption on the history-preference relation.

Histories and sequences of observations of length or dimension 2 are closely
parallel to the not necessarily symmetric case of ~\cite{LMS:98}. We first
recall the corresponding representation result in Section 3.2. Then, we show
that a simple generalization of this result fails already in the case of length
3 (Section 3.4).
Finally, in Section 3.5, we prove a valid representation theorem for the
general, $n$-dimensional case.

\subsection{The 2-D Representation Theorem}

First, let us quote a theorem characterizing the revision operators
representable by a pseudo-distance function (Proposition 2.5 of
~\cite{LMS:98}). The pseudo-distance mentioned here is actually no more than a
preference relation over pairs of models (or histories of length 2). The
theorem deals with an operator $\mid$ which revises a belief set by an
observation, i.e., \mbox{$A \mid B$} is the belief set held by an agent who has
held a belief set $A$, after observing $B$. Now, let $X$ be a finite and 
complete universe
(the set of possible models of the language). In such a universe, $A$ and
$B$ are interchangeably formulas, theories and sets of models. Let \mbox{$\cP
(X)$} designate the set of all {\em non-empty} subsets of $X$.

\begin{definition}
\label{def:rep2}
An operation
\mbox{$\mid \: : \cP ( X ) \times \cP ( X ) \rightarrow \cP ( X )$}
is {\em representable} iff there is a pseudo-distance
\mbox{$d : X \times X \rightarrow \cZ$} such that
\[
A \mid B = \{ b \in B \mid \exists a \in A {\rm \ such \ that \ }
\forall a' \in A , b' \in B , d(a, b) \leq d(a', b')\}.
\]
\end{definition}

Thus, intuitively, if $A$ and $B$ are sets, $A \mid B$ is the set of those
elements of $B$, which are closest to the set A. By abuse of notation, if $A$
and
$B$ are formulas, $A \mid B$ is the set of formulas valid in the set of those
models of $B$, which are closest to the set of models of $A$.

For this theorem, Lehmann, Magidor and Schlechta define a relation $R_{\mid}$
on pairs from $\cP ( X ) \times \cP ( X )$, which intuitively means ``provably
closer'' or ``provably preferable'', i.e., assuming the underlying
pseudo-distance exists, this relation represents information about it that may
be deduced by examining the revision operation. For instance, if
$(A \mid (B \xcv C)) \xcs B \xEd \xCQ$, $B$ is provably at least as close
to $A$, as $C$ is to $A$.
This information can only apply
to the best-preferred pairs of models in the pairs of sets, so
\mbox{$(A,B)R_{\mid}(A',B')$} actually means we have evidence that the best
pair of models in \mbox{$A \times B$} is at least as preferable as the best
pair in \mbox{$A' \times B'$}.

\begin{definition}
\label{def:R2}
Given an operation $\mid$ , define a relation $R_{\mid}$
on pairs from \cP $( X ) \times \cP ( X )$ by:
\mbox{$(A,B) R_{\mid} (A',B')$} iff one of the following two cases
obtains:

\begin{enumerate}
\item \label{right2}
\mbox{$A = A'$} and \mbox{$(A \mid (B \cup B')) \cap B \neq \emptyset$},
\item \label{left2}
\mbox{$B = B'$} and \mbox{$((A \cup A') \mid B) \neq (A' \mid B)$}.
\end{enumerate}

\end{definition}

In the rest of this subsection we shall write $R$ instead of $R_{\mid}$.
As usual, we shall denote by $R^{\star}$ the transitive closure of $R$.

Now, we can quote the representation theorem:

\begin{theorem}
\label{the:char2}
An operation $\mid$ is representable iff it satisfies the four
conditions below for any non-empty sets \mbox{$A, A', B, B' \subseteq X$}:

\begin{enumerate}
\item \label{cond:inc2}
\mbox{$(A \mid B) \subseteq B$},
\item \label{cond:leftor2}
\mbox{$((A \cup A') \mid B) \subseteq (A \mid B) \cup (A' \mid B)$},
\item \label{cond:right2}
If \mbox{$(A,B) R^{\star} (A,B')$}, then
\mbox{$(A \mid B) \subseteq (A \mid (B \cup B'))$},
\item \label{cond:left2}
If \mbox{$(A,B) R^{\star} (A',B)$}, then
\mbox{$(A \mid B) \subseteq ((A \cup A') \mid B)$}
\end{enumerate}

\end{theorem}

This is the strongest version of this theorem proven. Fixing the conditions
of the theorem, the characterization grows stronger as the definition of $R$
becomes narrower, as it then (possibly) puts less constraints on $\mid$. A
weaker characterization (which is also valid) has $R$ defined to be wider, as
follows:

\begin{definition}
\label{def:R2a}
We say the relation $R_{\mid}$ holds iff at least one of the following cases
obtains:

\begin{enumerate}
\item \label{inc2a} $A \supseteq A', B \supseteq B' \Rightarrow
(A,B)R(A',B')$
\item \label{right2a} $(A \mid (B \cup B')) \cap B \neq \emptyset \Rightarrow
(A,B)R(A,(B \cup B'))$
\item \label{left2a} $((A \cup A') \mid B) \neq (A' \mid B)\Rightarrow
(A,B)R((A \cup A'),B)$
\end{enumerate}

\end{definition}

\subsection{Ultimate Goal: The $n$-Dimensional Case}

We want to prove a theorem analogous to Theorem~\ref{the:char2} that
relates to strings of observations of length n (instead of length 2, if we
ignore the difference in role between a previous observation and a previous
belief set), that is, we want to characterize representable operations
\mbox{$[\;] \: : \cP ( X )^{n} \rightarrow \cP ( X )$}:

\begin{definition}
\label{def:rep}
An operation \mbox{$[\;] \: : \cP ( X )^{n} \rightarrow \cP ( X )$} is {\em
representable} iff there is a totally ordered set $\cZ$ (the order is $<$)
and a function
\mbox{$r :  X^{n} \rightarrow \cZ$} (that will be intuitively understood
as a history ranking),
such that, for any non-empty subsets \mbox{$A_{1} , ... A_{n} \subseteq X$},

\begin{equation}
\label{eq:rep}
\begin{array}{l}
[ A_1 \cdots A_n ] = \\
\ \{ a_n \in A_n \mid \exists a_1 \in A_1 ... a_{n-1} \in A_{n-1}
\forall a'_1 \in A_1 ... a'_n \in A_{n}
\: r(a_1,...,a_n) \leq r(a'_1,...,a'_n)\}
\end{array}
\end{equation}

\end{definition}

We will now see that this may not be achieved by straightforward
generalization of the tight 2-dimensional characterization, even for just three
dimensions.

\subsection{Simple Generalization is not Valid}

The first attempt at generalizing this theorem is held short at \mbox{$n=3$}.
Let us phrase the suggested theorem and disprove it, starting with a new
definition for $R$:

\begin{definition}
\label{def:R3}
Given an operation $[\;]$ , one defines a relation $R_{[\;]}$
on triplets of non-empty subsets of $X$ by:
\mbox{$(A , B , C) R_{[\;]} (A' , B' , C')$} iff one of the following cases
obtains:

\begin{enumerate}
\item \label{right3}
\mbox{$A = A'$}, \mbox{$B = B'$} and
\mbox{$[A \cdot B \cdot (C \cup C')] \cap C \neq \emptyset$}.
\item \label{middle3}
\mbox{$A = A'$}, \mbox{$C = C'$} and
\mbox{$[A \cdot (B \cup B') \cdot  C] \neq [A \cdot B' \cdot C] $}.
\item \label{left3}
\mbox{$B = B'$}, \mbox{$C = C'$} and
\mbox{$[(A \cup A') \cdot B \cdot C] \neq [A' \cdot B \cdot C] $}.
\end{enumerate}

\end{definition}

>From here on, unless otherwise stated, $R$ stands for $R_{[\;]}$. Now, the
suggested theorem:

\begin{suggest}
\label{sug:char}
An operation $[\;]$ is representable iff it satisfies the six
conditions below for any non-empty sets \mbox{$A,A',B,B',C,C' \subseteq X$}:

\begin{enumerate}
\item \label{cond:inc3}
\mbox{$[A \cdot B \cdot C] \subseteq C$}.
\item \label{cond:middleor3}
\mbox{$[A \cdot (B \cup B') \cdot C] \subseteq [A \cdot B \cdot C] \cup
[A \cdot B' \cdot C]$}.
\item \label{cond:leftor3}
\mbox{$[(A \cup A') \cdot B \cdot C] \subseteq [A \cdot B \cdot C] \cup
[A' \cdot B \cdot C]$}.
\item \label{cond:R3a}
\mbox{$(A,B,C)R^*(A',B,C) \Rightarrow [A \cdot B \cdot C] \subseteq
[(A \cup A') \cdot B \cdot C]$}.
\item \label{cond:R3b}
\mbox{$(A,B,C)R^*(A,B',C) \Rightarrow [A \cdot B \cdot C] \subseteq
[A \cdot (B \cup B') \cdot C]$}.
\item \label{cond:R3c}
\mbox{$(A,B,C)R^*(A,B,C') \Rightarrow [A \cdot B \cdot C] \subseteq
[A \cdot B \cdot (C \cup C')]$}.
\end{enumerate}

\end{suggest}

This theorem is not valid.

\proof
There is a counter-example, as follows:
Let \mbox{$X = \{0,1\}$} and \mbox{$n = 3$}. This seems to be the simplest
$n$-dimensional case for $n>2$. For convenience, we will write $0$ for
$\{0\}$, $1$ for $\{1\}$, $X$ for $\{0,1\}$, and $\star$ for any of them.
Define $[\;]$ as follows:
\[ [\star \cdot \star \cdot 0] = 0\ ;\  [\star \cdot \star \cdot 1] = 1 ; \]
\[ [0 \cdot 0 \cdot X] = 0\ ;\  [0 \cdot 1 \cdot X] = 0 ; \]
\[ [1 \cdot 0 \cdot X] = 1\ ;\  [1 \cdot 1 \cdot X] = X ; \]
\[ [0 \cdot X \cdot X] = 0\ ;\  [1 \cdot X \cdot X] = 1 ; \]
\[ [X \cdot 0 \cdot X] = 0\ ;\  [X \cdot 1 \cdot X] = X ; \]
\[ [X \cdot X \cdot X] = X. \]

The cases where $C=\{0\}$ and $C=\{1\}$ are forced by condition
\ref{cond:inc3} of the theorem, and are not very interesting. As for the case
$C=X$, one may check for each two triplets that fall under the conditions of
the theorem that it holds. The check is simplified by the fact that
Definition~\ref{def:R3} gives no way to show that \mbox{$(A,B,C)R(A',B',C)$}
when \mbox{$A \subseteq A'$} and \mbox{$B \subseteq B'$}, and the fact that
there is no valid union of sets of sequences of different cardinalities,
unless one is contained within the other. The operator complies with all the
conditions of Suggested Theorem~\ref{sug:char}.
But suppose there is
a ranking that defines $[\;]$, we reach the following conclusions (using
\mbox{$x \preceq y$} for \mbox{$x  R  y$}, and
\mbox{$x \prec y$} for \mbox{$x R y$} but provably \mbox{$\neg
y R^* x$}, and remembering that the conditions of the Suggested Theorem hold):
\[ (1 \cdot 1 \cdot X) \preceq (0 \cdot 1 \cdot X) \] by
Definition~\ref{def:R3}, case~\ref{left3},
\[ (1 \cdot 0 \cdot X) \prec (1 \cdot 1 \cdot X) \] by case~\ref{middle3} of
Definition~\ref{def:R3}, and negation of condition~\ref{cond:R3b} of the
Suggested Theorem, and
\[ (0 \cdot 0 \cdot X) \prec (1 \cdot 0 \cdot X) \] by case~\ref{left3} of the
definition of $R$ and negation of condition~\ref{cond:R3a} of the Suggested
Theorem.

Thus, the best history in \mbox{$\{0\} \cdot \{0\} \cdot X$} (which we know
to be \mbox{$ 0 \cdot 0 \cdot 0 $}) would also be the single best history in
\mbox{$X \cdot X \cdot X$}, but we find \mbox{$[X \cdot X \cdot X]=X$}
instead, which means there really is no underlying ranking. \QED

The next natural step is to use a definition more like Definition~\ref{def:R2a}
in the generalized theorem. If $R$ was defined as follows:

\begin{definition}
\label{def:R3a}
We say the relation $R_{[\;]}$ holds iff any of the following cases obtains:

\begin{enumerate}
\item $A \supseteq A', B \supseteq B', C \supseteq C' \Rightarrow
(A,B,C)R(A',B',C')$
\item $[A \cdot B \cdot (C \cup C')] \cap C \neq \emptyset \Rightarrow
(A,B,C)R(A,B,C \cup C')$
\item $[A \cdot (B \cup B') \cdot C] \neq [A \cdot B' \cdot C]\Rightarrow
(A,B,C)R(A,B\cup B',C)$
\item $[(A \cup A')\cdot B  \cdot C] \neq [A' \cdot B \cdot C]\Rightarrow
(A,B,C)R(A\cup A',B,C)$
\end{enumerate}

\end{definition}

Then, for the operator defined in the counter-example, we could show that
\[(1,0,X)R(1,X,X)R(X,X,X)R(X,0,X)\] but \[[1 \cdot 0 \cdot X] \not \subseteq
[X \cdot 0 \cdot X]\] so the operator violates condition~\ref{cond:R3a}, and is
thus not a counter-example. We have not been able to prove a representation
theorem using Definition~\ref{def:R3a}, and believe it not to be valid. Yet,
we have not found any counter-example. We have discovered (by computerized
enumeration of the operators) that there are no counter-examples with
\mbox{$n=3$} and \mbox{$\lmid X \rmid = 2$}, and the question whether it is
valid for all $n$ and \mbox{$\lmid X \rmid$}, and if not, for which $n$ and
\mbox{$\lmid X \rmid$} it is valid, remains open at this stage.

\subsection{An $n$-Dimensional Representation Theorem}

\subsubsection{Patches}

We believe that for this theorem, we need a way to work around ``illegal" sets
of histories, i.e. such that are not sequences of sets of models. We call such
sets ``illegal" because, not being interchangeable with sequences of
observations, our operator cannot be applied to them. For the technique used in
this proof, it is especially regretful that for most sequences $\sigma$, if we
try to present $\sigma$ as a disjoint union of a ``legal" sequence (say, a
singleton) $\tau$ with the rest of the histories in the sequence $\sigma
\setminus \tau$ then the latter will be illegal. Instead, we use a family of
legal sequences whose union is $\sigma \setminus \tau$. As it complements
$\tau$ to $\sigma$, we call this family a {\em patch}.

\begin{definition}
\label{def:patch}
Given a sequence \mbox{$A_1 \cdots A_n$} and a sequence \mbox{$A'_1 \cdots
A'_n$} such that \mbox{$\forall i A'_i \subseteq A_i$}, we define the {\em
patch to} \mbox{$A_1 \cdots A_n$} {\em from} \mbox{$A'_1 \cdots A'_n$} to be
the set of sequences comprised of all the sequences \mbox{$B^i_1 \cdots
B^i_n$} for each $i$ such that \mbox{$ A'_i \neq A_i$}, where

\begin{enumerate}
\item $\forall j \neq i$, $B^i_j = A_j$
\item $B^i_i = A_i \setminus A'_i$
\end{enumerate}

\end{definition}

Let, for instance, $n=3$, $A:=\{0,1\} \cdot \{0,1\} \cdot \{0,1\}$,
$A':=\{0\} \cdot \{0,1\} \cdot \{0\}$, then the patch to $A$ from $A'$ is
$\{B^1:=\{1\} \cdot \{0,1\} \cdot \{0,1\},
B^3:=\{0,1\} \cdot \{0,1\} \cdot \{1\} \}$.

Note that the (disjoint)
union of \mbox{$A'$} with \mbox{$\bigcup_iB^i$} is
equal to $A$.

As with the definition of $R_{\mid}$, this definition is made as tight as
possible to make the theorem stronger. We will see later that if we relax the
definition a little, allowing many patches from $A'$ to $A$ by letting
$B^i_i$ be any set such that \mbox{$A'_i \cup B^i_i = A_i$}, our theorem will
 become the generalization of the weaker form of Theorem~\ref{the:char2} (the
one using Definition~\ref{def:R2a}). Notice also that
Definition~\ref{def:patch} may be
easily written in the language of formulae; we do not wish our basic
concepts, or the phrasing of the representation theorem, to stray too far
from our original problem domain.

\subsubsection{The Theorem}

With the new tool, the patch, we may define a new, more powerful relation of
``provable preferability".

\begin{definition}
\label{def:R}
Given an operation $[\;]$, define a relation $R_{[\;]}$ on sequences of
length $n$ of non-empty subsets of $X$ by:
\mbox{$(A'_1 ,..., A'_{n-1} , C') R_{[\;]} (A_1 ,..., A_{n-1} , C)$} iff one of
the following cases
obtains:

\begin{enumerate}
\item \label{inclusion}
\mbox{$\forall i A_i \subseteq A'_i$} and \mbox{$C \subseteq C'$}.
\item \label{right}
\mbox{$\forall i A_i = A'_i$}, \mbox{$C' \xcc C$} and
\mbox{$[A_1 \cdots A_{n-1} \cdot C] \cap C' \neq \emptyset$}.
\item \label{left}
\mbox{$\forall i A_i \supseteq A'_i$}, \mbox{$C = C'$},
\mbox{$\{B^i_1 \cdots B^i_{n-1}\}$} is the patch to \mbox{$A_1 \cdots A_{n-
1}$} from \mbox{$A'_1 \cdots A'_{n-1}$}, and one of the following holds:

\begin{enumerate}
\item \label{intersection}
\mbox{$\bigcap_{i} [B^i_1 \cdots B^i_{n-1} \cdot C] \not \subseteq [A_1
\cdots A_{n-1} \cdot C]$}
\item \label{union}
\mbox{$[A_1 \cdots A_{n-1} \cdot C] \not \subseteq \bigcup_{i} [B^i_1 \cdots
B^i_{n-1} \cdot C]$}
\end{enumerate}

\end{enumerate}

\end{definition}

Now we are ready to phrase the
representation theorem for the $n$-dimensional case:

\begin{theorem}
\label{the:char}
An operation $[\;]$ is representable iff it satisfies the three
conditions below for any non-empty sets \mbox{$A_1 ,... A_{n-1} , C \subseteq
X$}:

\begin{enumerate}
\item \label{cond:inc}
\mbox{$[A_1 \cdots A_{n-1} \cdot C] \subseteq C$},
\item \label{cond:leftor}
If \mbox{$\forall i A'_i \subseteq A_i$} and \mbox{$\{B^i_1 \cdots B^i_{n-
1}\}$} is the patch to \mbox{$A_1 \cdots A_{n-1}$} from \mbox{$A'_1 \cdots
A'_{n-1}$} then \mbox{$[A_1 \cdots A_{n-1} \cdot C] \subseteq [A'_1 \cdots
A'_{n-1} \cdot C] \cup \bigcup_{i} [B^i_1 \cdots B^i_{n-1} \cdot C]$}
\item \label{cond:R}
If \mbox{$\forall i A'_i \subseteq A_i$}, \mbox{$C' \subseteq C$} and
\mbox{$(A'_1 \cdots A'_{n-1} \cdot C') R^{\star} (A_1 \cdots A_{n-1} \cdot
C)$}, then \\
\mbox{$[A'_1 \cdots A'_{n-1} \cdot C'] \subseteq [A_1 \cdots A_{n-1} \cdot
C]$}.
\end{enumerate}

\end{theorem}

Before proving this theorem, we note that when taking the more relaxed
version (as mentioned above) of Definition~\ref{def:patch}, and taking
Definition~\ref{def:R2a} to define $R_{\mid}$, Theorem~\ref{the:char2}
becomes the special case of Theorem~\ref{the:char} with $n=2$: Under these
definitions, for any sets $A$ and $A'$, $A'$ (or rather, the set comprised of
it as a single sequence of length 1) is a patch to \mbox{$A \cup A'$}
from $A$, so the concept of the patch coincides with union. When the patch
has only one sequence (of one set) in it, then \mbox{$\bigcap [B^1_1 \cdot C] =
\bigcup [B^1_1 \cdot C] = [B^1_1 \cdot C]$}, and so the
cases~\ref{intersection} and~\ref{union} of
Definition~\ref{def:R} together become case~\ref{left2a} of
Definition~\ref{def:R2a}. And finally, the conditions~\ref{cond:right2}
and~\ref{cond:left2} of Theorem~\ref{the:char2} are expressed together as
condition~\ref{cond:R} of Theorem~\ref{the:char}.
Note that condition 3 in Theorem~\ref{the:char} is essentially a loop 
condition.
Let us now prove the more general theorem.

\proof
For the proof of Theorem~\ref{the:char}, a number of lemmas will be needed.
These lemmas will be presented when needed, and their proof inserted in the
midst of the main proof. First, we shall deal with the {\em soundness} of the
theorem, and then with the more challenging {\em completeness}.\\

Suppose, then, that \mbox{$[\;]$} is representable. The ranking $r$ of
histories may be extended to sequences of sets in the usual way, by taking
the minimum over the sets: \mbox{$r(A_1 ,..., A_n) = min_{a_i \in A_i} \{
r(a_1,..., a_n)\}$}. One may then write the equation defining representability
as
\[
[A_1 \cdots A_{n-1} \cdot C] = \{ c \in C \mid r(A_1 ,..., A_{n-1},\{c\}) =
r(A_1 ,..., A_{n-1},C) \}
\]
Let us now show that the three conditions of Theorem~\ref{the:char} hold:\\
Condition~\ref{cond:inc} is obvious. For Condition~\ref{cond:leftor}, notice
that, on one hand, \linebreak \mbox{$A_1 \cdots A_{n-1}= \bigcup_{i} \{B^i_1
\cdots B^i_{n-1}\} \cup (A'_1 \cdots A'_{n-1})$}, and on the other hand,
\linebreak if \mbox{$r(a_1,..., a_{n-1},c) = r(A_1 ,..., A_{n-1},C)$},
\mbox{$\langle a_1, \cdots , a_n, c \rangle \in A''_1 \cdots A''_{n-1} \cdot
C''$}
for some $A_i'', C''$, and
\mbox{$ A''_1 \cdots A''_{n-1} \cdot C'' \subseteq A_1 \cdots
A_{n-1}\cdot C$} then also \mbox{$r(a_1,..., a_{n-1},c) = r(A''_1 ,..., A''_{n-
1},C'')$} and \mbox{$c \in [A''_1 \cdots A''_{n-1} \cdot C'']$}. For
Condition~\ref{cond:R} we need a little lemma:

\begin{lemma}
\label{lem:R-imp-leq}
For any non-empty sets \mbox{$A_i, A'_i, C, C' \subseteq X$},
\[
(A'_1,..., A'_{n-1},C')R(A_1,..., A_{n-1},C) \Rightarrow r(A'_1,..., A'_{n-
1},C')\leq r(A_1,..., A_{n-1},C)
\]
\end{lemma}

\proof
Let us consider the different cases of Definition~\ref{def:R}:\\
Case~\ref{inclusion} is obvious.\\
In case~\ref{right} the negation of the consequence \mbox{$r(A_1,..., A_{n-
1},C)<r(A'_1,..., A'_{n-1},C')$} implies \mbox{$[A_1 \cdots A_{n-1} \cdot C]
\cap C' = \emptyset$} which is the negation of the assumption.\\
Both sub-cases of case~\ref{left} rely on the reasoning used for
Condition~\ref{cond:leftor} above:
In case~\ref{intersection} by assumption there exists $c$ such that \mbox{$c
\in \bigcap_{i} [B^i_1 \cdots B^i_{n-1} \cdot C]$} but \mbox{$c \not \in
[A_1 \cdots A_{n-1} \cdot C]$}. By the former, for each $i$ there is a
history
\mbox{$\langle b^i_1, \cdots , b^i_{n-1}, c \rangle$},
\mbox{$b^i_j \in B^i_j$},
with $r(b^i_1, \cdots, b^i_{n-1} , c)$ minimal
in \mbox{$ B^i_1 \cdots B^i_{n-1} \cdot C$}, but by the
latter, for none of these histories $r(b^i_1, \cdots , b^i_{n-1}, c)$ is
minimal in \mbox{$A_1 \cdots A_{n-1}
\cdot C$}. So the histories with $r$ minimal in \mbox{$A_1 \cdots A_{n-1}
\cdot C$} have to be all members of \mbox{$A'_1 \cdots A'_{n-1} \cdot C$}
which implies \mbox{$r(A'_1,..., A'_{n-1},C)\leq r(A_1,..., A_{n-1},C)$} as
needed.\\
In case~\ref{union}, by assumption there is at least one history that is one
of the best in \mbox{$A_1 \cdots A_{n-1} \cdot C$}, but is not one of the
best in (and thus not a member of) \mbox{$ B^i_1 \cdots B^i_{n-1} \cdot C$}
for any $i$. Hence this history is in \mbox{$A'_1 \cdots A'_{n-1} \cdot C$}
and \mbox{$r(A'_1,..., A'_{n-1},C)\leq r(A_1,..., A_{n-1},C)$} as needed.\QED
We conclude that Condition~\ref{cond:R} holds and the characterization is
sound.\\
For the completeness part, assume the operator $[\;]$ complies with the
conditions of Theorem~\ref{the:char}. Using the Generalized Abstract Nonsense
Lemma 2.1 of ~\cite{LMS:98}, extend $R$ to a total
preorder $S$  satisfying

\begin{equation}
\label{eq:r-and-s}
x S y , y S x \Rightarrow x R^{\star} y.
\end{equation}

Let $\cZ$ be the totally ordered set of equivalence classes of
\mbox{$\cP(X)^n $} defined by the total pre-order $S$.
Define a function \mbox{$d\ :\ \cP(X)^n \rightarrow \cZ$} to send a sequence
of subsets $A_1,..., A_n$ to its equivalence class under $S$.
We shall define \mbox{$r\ :\ X^n \rightarrow \cZ$} by
\mbox{$r(a_1,...,a_n)\eqdef d(\{a_1\},...,\{a_n\})$}. While we aim to prove
that $r$ represents $[\;]$, we will have to use $d$ for the proof. $d$ has
the following obvious properties:

\begin{equation}
\label{eq:R-imp-d}
\begin{array}{l}
(A_1,..., A_{n-1},C) R (A'_1,..., A'_{n-1},C') \\
\ \Rightarrow d(A_1,..., A_{n-1},C) \leq d(A'_1,..., A'_{n-1},C'),
\end{array}
\end{equation}

and, from Equation~\ref{eq:r-and-s},

\begin{equation}
\label{eq:d-and-s}
\begin{array}{l}
 d(A_1,..., A_{n-1},C) = d(A'_1,..., A'_{n-1},C') \\
\ \Rightarrow (A_1,..., A_{n-1},C)R^\star(A'_1,..., A'_{n-1},C').
\end{array}
\end{equation}

$d$ also has the less obvious property shown by the next lemma, which means it
approximates representation as far as the last argument is concerned:

\begin{lemma}
\label{le:partial}
For any $A_1,..., A_{n-1},C$, \[d(A_1,..., A_{n-1},C) = min_{c \in C}
\{d(A_1,..., A_{n-1},\{c\})\}\] and

\begin{equation}
\label{eq:part}
[A_1 \cdots A_{n-1} \cdot C] = \{ c \in C \mid d(A_1,..., A_{n-1}, \{c\}) =
d(A_1,..., A_{n-1}, C)\}.
\end{equation}

\end{lemma}

\proof
Suppose \mbox{$c \in C$}, then \mbox{$(A_1,..., A_{n-1}, C) R (A_1,..., A_{n-
1}, \{ c \})$} and
from Equation~\ref{eq:R-imp-d} we get \mbox{$d(A_1,..., A_{n-1},C) \leq min_{c
\in C} \{d(A_1,..., A_{n-1},c)\}$}. If moreover \mbox{$c \in [A_1 \cdots A_{n-
1} \cdot C]$}, then \mbox{$[A_1 \cdots A_{n-1} \cdot C] \cap \{c\} \neq 
\emptyset$}, and by Definition~\ref{def:R}, part~\ref{right},
\mbox{$(A_1,..., A_{n-1}, \{c\}) R (A_1,..., A_{n-1}, C)$} and therefore
\linebreak \mbox{$d(A_1,..., A_{n-1}, c) = d(A_1,..., A_{n-1}, C)$}. We have
shown that the left hand side of Equation~\ref{eq:part} is included in the
right hand side.\\
Since \mbox{$[A_1 \cdots A_{n-1} \cdot C]$} is not empty,
\mbox{$\exists c \in [A_1 \cdots A_{n-1} \cdot C]$} and, by the previous
remark,
\mbox{$d(A_1,..., A_{n-1}, C) = d(A_1,..., A_{n-1}, c)$} and therefore we
conclude that
\mbox{$d(A_1,..., A_{n-1},C) = min_{c \in C} \{d(A_1,..., A_{n-1},c)\}$}.

To see the converse inclusion, notice that
\mbox{$d(A_1,..., A_{n-1}, C) = d(A_1,..., A_{n-1}, c)$} implies
\mbox{$(A_1,..., A_{n-1}, c) R^{\star} (A_1,..., A_{n-1}, C)$} and, by
Property~\ref{cond:R} of Theorem~\ref{the:char},
\mbox{$[A_1 \cdots A_{n-1} \cdot \{c\}] \subseteq [A_1
\cdots A_{n-1} \cdot C]$}, so \mbox{$c \in [A_1 \cdots A_{n-1} \cdot C]$} by
Property~\ref{cond:inc} of the theorem. \QED
We now have to show that

\begin{equation}
\label{eq:rep3}
\begin{array}{l}
[A_1 \cdots A_{n-1} \cdot C] = \\
\ \{ c \in C \mid \exists a_1,..., a_{n-1}, \ s.t. \ \forall a'_1,..., a'_{n-
1}, c' \in C,\ r(a_1,..., a_{n-1},c) \leq r(a'_1,..., a'_{n-1},c') \}
\end{array}
\end{equation}

(where \mbox{$\forall i \ a_i,a'_i \in A_i$}).

To see that the right hand side
is a subset of the left hand side, assume that \mbox{$a_i \in A_i, c \in C$}
are such that for all \mbox{$a'_i \in A_i, c' \in C$},
\mbox{$r(a_1,..., a_{n-1},c) \leq r(a'_1,..., a'_{n-1},c')$}. We have to show
that \mbox{$c \in [A_1 \cdots A_{n-1} \cdot C]$}.
We will show by induction
on the size of \mbox{$A'_1 \cdots A'_{n-1}$}, where
 \mbox{$ a_i \in A'_i \subseteq A_i$} for all $i$,
that \mbox{$c \in [A'_1 \cdots A'_{n-1} \cdot C]$}.

For the base of the induction, if \mbox{$A'_1 \cdots A'_{n-1}$} is a
singleton, then \mbox{$A'_i=\{a_i\}$} and by Lemma~\ref{le:partial},
remembering that in this case $r$ coincides with $d$, we find \mbox{$c \in
[A'_1 \cdots A'_{n-1} \cdot C]$}.\\
Otherwise, i.e. if not all $A_i'$ are singletons,
we may choose for each $i$ an \mbox{$a'_i \in A'_i$} such that
\mbox{$a'_i \neq a_i$} if \mbox{$A'_i \neq \{a_i\}$}. Since \mbox{$A'_1
\cdots A'_{n-1}$} is not a singleton, at least one of the set inequalities
holds and \mbox{$ \langle a'_1, \cdots , a'_{n-1} \rangle \neq \langle a_1,
\cdots , a_{n-1} \rangle $}.\\
Let \mbox{$\{B^i_1 \cdots B^i_{n-1}\}$} be the patch to \mbox{$A'_1 \cdots
A'_{n-1}$} from \mbox{$\{a'_1\} \cdots \{a'_{n-1}\}$}. For all $i$ we have,
by definition of the patch, \mbox{$ \lmid B^i_1 \cdots B^i_{n-1} \rmid <
\lmid A'_1 \cdots A'_{n-1} \rmid $}. By choice of $a'_i$, \mbox{$\langle a_1,
\cdots ,
a_{n-1} \rangle \in B^i_1 \cdots B^i_{n-1}$} for all $i$ and hence by the
induction hypothesis,
\mbox{$c \in [B^i_1 \cdots B^i_{n-1} \cdot C]$}. Assume now that \mbox{$c
\not \in [A'_1 \cdots A'_{n-1} \cdot C]$}, then by
Property~\ref{intersection} of Definition~\ref{def:R}, \mbox{$(a'_1,...,
a'_{n-1},C) R (A'_1,..., A'_{n-1},C)$}. There is some \mbox{$c' \in [a'_1
\cdots a'_{n-1} \cdot C]$} and by Definition~\ref{def:R} part~\ref{right},
\mbox{$(a'_1,..., a'_{n-1},c') R (a'_1,..., a'_{n-1},C)$}. Also, by
Definition~\ref{def:R} part~\ref{inclusion}, \mbox{$(A'_1,..., A'_{n-1},C) R
(a_1,..., a_{n-1},c)$}. We have established that \mbox{$(a'_1,..., a'_{n-1},c')
R^{\star} (a_1,..., a_{n-1},c)$} so by our original assumption we have
\mbox{$d(a'_1,..., a'_{n-1},c') = d(a_1,..., a_{n-1},c)$}. By
Equation~\ref{eq:d-and-s} this implies also \mbox{$(a_1,..., a_{n-
1},c) R^{\star} (a'_1,..., a'_{n-1},c')$} and hence \mbox{$(a_1,..., a_{n-1},c)
R^{\star} (A'_1,..., A'_{n-1},C)$}. But then by Conditions \ref{cond:R} and
\ref{cond:inc} of Theorem~\ref{the:char}, we get \mbox{$c \in [A'_1 \cdots
A'_{n-1} \cdot C]$}
which is a contradiction. We conclude that the right hand side of
Equation~\ref{eq:rep3} is a subset of the left hand side.

For the converse assume that \mbox{$c \in C$} is such that for all
\mbox{$a_1,..., a_{n-1}$} \mbox{$(a_i \in A_i), $} there exist \mbox{$a'_1,...,
a'_{n-1}$} and \mbox{$c' \in C$} such that $c \neq c'$ and
\mbox{$r(a_1,..., a_{n-1},c) \not
\leq r(a'_1,..., a'_{n-1},c')$}. We need to prove that \mbox{$c \not \in [A_1
\cdots A_{n-1} \cdot C]$}. Since $X$ is finite, we may change the order of
quantifiers in the assumption to
\[
\exists a'_1,..., a'_{n-1}, c' \in C, \forall a_1,..., a_{n-1}, \: r(a_1,...,
a_{n-1},c) \not \leq r(a'_1,..., a'_{n-1},c')
\]
Note that \mbox{$r(a_1,..., a_{n-1},c) \not \leq r(a'_1,..., a'_{n-1},c')
\Rightarrow \neg (a_1,..., a_{n-1},c) R^{\star}(a'_1,..., a'_{n-1},c')$}.\\
As above, but switching the roles of $a_i$ with those of
$a'_i$, let $A'_1,..., A'_{n-1}$ be such that \mbox{$ a'_i \in A'_i \subseteq
A_i$}, and prove by induction on \mbox{$\lmid A'_1 \cdots A'_{n-1}\rmid$},
i.e. the cardinality of \mbox{$ A'_1 \cdots A'_{n-1}$},
that \mbox{$c \not \in [A'_1 \cdots A'_{n-1} \cdot C]$}. If \mbox{$A'_1
\cdots A'_{n-1}$} is a singleton, then
\mbox{$A'_i=\{a'_i\}$}, and by Lemma~\ref{le:partial} \mbox{$c \not \in [A'_1
\cdots A'_{n-1} \cdot C]$}. Otherwise choose for each $i$ an $a_i \in A'_i$
with  $a_i \neq a'_i$ if possible, and let \mbox{$\{B^i_1 \cdots B^i_{n-
1}\}$} be the patch to \mbox{$A'_1 \cdots A'_{n-1}$} from \mbox{$\{a_1\}
\cdots \{a_{n-1}\}$}. Again we have for all $i$, \mbox{$\langle a'_1, \cdots ,
a'_{n-1} \rangle
\in B^i_1 \cdots B^i_{n-1}$} and \mbox{$ \lmid B^i_1 \cdots B^i_{n-1} \rmid <
\lmid A'_1 \cdots A'_{n-1} \rmid $} and hence by induction
\mbox{$c \not \in \bigcup_i[B^i_1 \cdots B^i_{n-1} \cdot C]$}.
Now if \mbox{$c \in [A'_1 \cdots A'_{n-1} \cdot C]$} we get two consequences.
First, by Definition~\ref{def:R}, part~\ref{union}, \mbox{$(a_1,..., a_{n-
1},C)R(A'_1,..., A'_{n-1},C)$}. By the same definition, part~\ref{inclusion},
we have \mbox{$(A'_1,..., A'_{n-1},C)R(a'_1,..., a'_{n-1},c')$}. Second, from
condition~\ref{cond:leftor} of Theorem~\ref{the:char}, we derive \mbox{$c \in
[a_1 \cdots a_{n-1} \cdot C]$} so that \mbox{$(a_1,..., a_{n-1},c)R(a_1,...,
a_{n-1},C)$}. The above together give us \mbox{$(a_1,..., a_{n-
1},c)R^{\star}(a'_1,..., a'_{n-1},c')$} which contradicts our assumption. We
have thus shown the converse inclusion. \QED

\section{Conclusion}
Iterated updates, of central importance for AI, cannot be properly
treated in the AGM framework, not because of the update vs. revision
distinction as was thought by Katsuno-Mendelzon, but because of the
identification of epistemic states with belief sets, identification
accepted by Katsuno-Mendelzon.
An ontology in which epistemic states are richer than
belief sets yields a family of updates that satisfy the AGM assumptions.
We have first proved in Section 2 several properties, which are intuitively
appealing, and extend the AGM postulates. In Section 3, we have given a
complete set of properties, which characterize our approach, and shown a
representation theorem (Theorem 3.2). In summary, we have - on the 
philosophical
side - introduced a natural and intuitively interesting ontology for iterated
update, and - on the mathematical side - characterized our idea with a set of
sound and complete conditions.

\section{Acknowledgements}

The authors would like to thank two anonymous referees who have helped to make
the article much clearer.


\begin{thebibliography}{1}

\bibitem{ADJP:96}
Adnan Darwiche and Judea Pearl.
\newblock On the Logic of Iterated Belief Revision.
\newblock Artificial Intelligence, 89(1-2):1--29, 1997.

\bibitem{NFJH:96}
Nir Friedman and Joseph Y. Halpern.
\newblock Belief Revision: A Critique.
\newblock Proceedings of the Fifth International Conference on Principles of
Knowledge Representation and Reasoning, KR'96.
\newblock Edited by Luigia Carlucci Aiello, Jon Doyle and Stuart Shapiro.
\newblock Morgan Kaufmann, Cambridge, Mass., November 1996
\newblock Pages 421--431.

\bibitem{Gard:intro}
Peter G\"{a}rdenfors.
\newblock Belief revision: An introduction.
\newblock In Peter G\"{a}rdenfors, editor, {\em Belief Revision}, number~29 in
  Cambridge Tracts in Theoretical Computer Science, pages 1--28. Cambridge
  University Press, 1992.

\bibitem{Gard:book}
Peter G\"{a}rdenfors.
\newblock Knowledge in Flux.
\newblock MIT Press, Cambridge, Massachusetts, 1988.

\bibitem{KatMend:92}
Hirofumi Katsuno and Alberto~O. Mendelzon.
\newblock On the difference between updating a knowledge base and revising it.
\newblock In Peter G\"{a}rdenfors, editor, {\em Belief Revision}, number~29 in
  Cambridge Tracts in Theoretical Computer Science, pages 183--203. Cambridge
  University Press, 1992.

\bibitem{Leh:IJCAI95}
Daniel Lehmann.
\newblock Belief revision, revised.
\newblock In {\em Proceedings of 14th IJCAI}, pages 1534--1541, Montreal,
  Canada, August 1995. Morgan Kaufmann.

\bibitem{LMS:98}
Daniel Lehmann, Menachem Magidor, Karl Schlechta.
\newblock Distance Semantics for Belief Revision.
\newblock Hebrew University Technical Report TR-98-10.
\newblock Department of Computer Science, Hebrew University,
Givat Ram, Jerusalem 91904, Israel.

\bibitem{Nebel}
Bernhard Nebel.
\newblock A Knowledge Level Analysis of Belief Revision.
\newblock In {\em Proceed. First International Conference on Principles of
Knowledge Representation and Reasoning}, Toronto, 1989.

\bibitem{SLM:96}
Karl Schlechta, Daniel Lehmann and Menachem Magidor.
\newblock Distance Semantics for Belief Revision.
\newblock Proceedings of the Sixth Conference on Theoretical Aspects of
Rationality and Knowledge
\newblock Edited by Yoav Shoham.
\newblock Morgan Kaufmann, De Zeeuwse Stromen, The Netherlands, March 1996
\newblock Pages 137--145.

\bibitem{Sch:95}
Karl Schlechta.
\newblock Preferential Choice Representation Theorems for Branching
Time Structures.
\newblock Journal of Logic and Computation, Oxford, Vol. 5, pp. 783-800, 1995.


\end{thebibliography}
\end{document}